\documentclass{article}

\PassOptionsToPackage{numbers,compress,sort}{natbib}
\usepackage[final]{nips_2016}

\usepackage[utf8]{inputenc} %
\usepackage[T1]{fontenc}    %
\usepackage{hyperref}       %
\usepackage{url}            %
\usepackage{booktabs}       %
\usepackage{amsfonts}       %
\usepackage{nicefrac}       %
\usepackage{microtype}      %

\usepackage{amsmath}
\usepackage{mathtools,xparse}
\usepackage[pdftex]{graphicx}

\hypersetup{pdfauthor={Artem Babenko},pdftitle={Pairwise Quantization}}

\newcommand{\fig}[1]{Figure~\ref{fig:#1}}
\newcommand{\sect}[1]{Section~\ref{sect:#1}}
\newcommand{\tab}[1]{Table~\ref{tab:#1}}

\newcommand{\eq}[1]{(\ref{eq:#1})}

\renewcommand{\paragraph}[1]{\smallskip\noindent{\bf{#1}}}

\DeclareMathOperator*{\argmin}{\arg\!\min}
\newcommand{\E}[1]{\mathrm{\mathbb E}\left(#1\right)}
\newcommand{\Ek}[1]{\mathrm{\mathbb E_k}\left(#1\right)}

\newcommand{\sqn}[1]{{\left\|#1\right\|^2}}
\newcommand{\sqd}[2]{\sqn{#1-#2}}
\newcommand{\spr}[2]{{\langle{}#1,#2\rangle}}

\newcommand{\Q}{Q}
\newcommand{\X}{X}

\title{Pairwise Quantization}

\author{
  Artem Babenko \\
  Yandex\\
  Moscow Institute of Physics and Technology\\
  \texttt{artem.babenko@phystech.edu} \\
  \And
  Relja Arandjelovi\'c\thanks{%
      {\scriptsize WILLOW project,
      Departement d'Informatique de l'\'Ecole Normale Sup\'erieure,
      ENS/INRIA/CNRS UMR 8548.}%
      } \\
  WILLOW Project \\
  INRIA \\
  \texttt{relja.arandjelovic@inria.fr} \\
  \And
  Victor Lempitsky \\
  Skolkovo Institute of Science and Technology \\
  (Skoltech) \\
  \texttt{lempitsky@skoltech.ru} \\
  \\
}

\begin{document}

\maketitle

\begin{abstract}
We consider the task of lossy compression of high-dimensional vectors through quantization. We propose the approach that learns quantization parameters by minimizing the distortion of scalar products and squared distances between \textit{pairs} of points. This is in contrast to previous works that obtain these parameters through the minimization of the reconstruction error of individual points. The proposed approach proceeds by finding a linear transformation of the data that effectively reduces the minimization of the pairwise distortions to the minimization of individual reconstruction errors. After such transformation, any of the previously-proposed quantization approaches can be used. Despite the simplicity of this transformation, the experiments demonstrate that it achieves considerable reduction of the pairwise distortions compared to applying quantization directly to the untransformed data.
\end{abstract}

\section{Introduction}

Approaches based on quantization with multiple codebooks (such as product quantization \cite{Jegou11a}, residual quantization \cite{Chen10}, optimized product quantization \cite{Ge13,Norouzi13}, additive quantization \cite{Babenko14}, composite quantization \cite{Zhang14}, tree quantization \cite{Babenko15}) perform lossy compression of high-dimensional vectors, achieving very good trade-off between accuracy at preserving the high-dimensional vectors, speed, and compression ratio.  In many scenarios associated with machine learning and information retrieval, one is interested to preserve the \textit{pairwise} relations between high-dimensional points (e.g.\ distances, scalar products) rather than the points themselves.
This is not reflected in the existing quantization approaches, which invariably learn their parameters (e.g.\ codebooks) from data by minimizing the reconstruction (compression) error of the training data points.  

In this work, we aim to derive quantization-based methods that directly minimize distortions of pairwise relations between points rather than reconstruction errors of individual points. In particular, we focus on minimizing the distortion of scalar products $\spr{q}{x}$ and squared Euclidean distances $\sqd{q}{x}$ between \textit{uncompressed} $q$ and \textit{compressed} $x$ that might come from two different distributions. This case is important for two reasons. Firstly, it is often beneficial (in terms of accuracy) to compress only one argument (e.g.\ when one needs to evaluate distances between a user query $q$ and a large dataset $x$ of points, there is no need to compress vectors expressing user queries). Secondly, quantization-based methods are particularly efficient at evaluating pairwise relations between compressed and uncompressed points as they can use the \textit{asymmetric distance computation} (ADC) scheme \cite{Jegou11a}. Given an uncompressed $q$ and a very large dataset of compressed vectors $x_1,\dots, x_N$, the ADC scheme precomputes the look-up tables containing the distances between $q$ and the codewords, and then does a linear scan through compressed vectors while estimating distances or scalar products by summing up values from the look-up tables.

Our approach proceeds by reducing the task of minimizing the pairwise relation distortions to the original task of minimizing the reconstruction error of individual points. We demonstrate that both for scalar products and for pairwise distances, such reduction can be achieved through simple linear transforms estimated from data. After such transforms, any of the existing quantization schemes \cite{Jegou11a,Chen10,Ge13,Norouzi13,Babenko14,Zhang14,Babenko15} can be applied. In the experiments, we demonstrate that applying the proposed transform allows to reduce the quantization-induced distortions of pairwise relations very considerably, as compared to applying quantization to the original vectors.

We discuss several practical applications where this improvement can matter for scalar products. For squared Euclidean distances, we find that our approach is dominated by large distances and hence is only beneficial when preserving long distances (rather than distances between close neighbors) matter.

\section{Review of Vector and Product Quantization}

\label{sect:opq}
Since the topic of this paper is vector quantization, and as our method relies
heavily on existing vector quantization approaches (\sect{pairq}), here we
provide a brief overview of a state-of-the-art approach called Optimized Product Quantization (OPQ) \cite{Ge13,Norouzi13}.

The goal of all existing vector quantization methods is to represent vectors well
while achieving large compression rates, i.e.\ for a given memory budget
and training vectors $\{x_i\}$, one wants to quantize them such that their
reconstructions $\{\hat{x}_i\}$ achieve the minimal quantization error:
\begin{equation}
L_q= \sum_i \sqd{x_i}{\hat{x_i}}
\label{eq:quantLoss}
\end{equation}
As will be shown in Sections~\ref{sect:method:dot} and~\ref{sect:method:sqd},
methods which minimize quantization distortions are suboptimal when it comes
to the task of achieving best scalar product or squared Euclidean distance
estimates.

A method which directly minimizes the quantization distortion is k-means. Thus,
a codebook $\{c_i\}$ of size $K$ is learnt using k-means and each vector is approximated
using its nearest codeword, i.e.\ the assignment of vector $x$ is done using
$a(x)=\argmin_k\sqd{x}{c_k}$ and the ID of the nearest codeword is used
for the compressed representation of the vector,
giving rise to a $\log_2 K$-bit representation.
The approximated version $\hat{x}$ of $x$ is then the closest codeword itself $\hat{x}=c_{a(x)}$.

Given a large dataset $\{x_i\}$, a memory-efficient way of computing
an approximate value of any scalar function $f(x)$ for all the vectors is:
(i) off-line stage: compress all database vectors and store their
IDs $\{a(x_i)\}$,
(ii) online stage: compute $f(c_k)$ for each $k$ in $1, 2, \dots, K$,
and then simply look up this value for each compressed database vector:
$f(x_i) \approx f(\hat{x}_i) = f(c_{a(x_i)})$.
For example, given a compressed database $\{a(x_i)\}$, such a system
is able to quickly estimate a scalar product between any query vector $q$
and the entire database by pre-computing all $q^T c_k$
and then looking up these values $q^Tx_i \approx q^T c_{a(x_i)}$.

However, this method is impractical in real life situations as for moderate
and large bitrates one would require prohibitively large codebooks, e.g.\ for
32 bits and 64 bits the codebook would contain 4 billion and $1.8 \times 10^{19}$
vectors, respectively. This is impractical as
(i) it is impossible to learn such a large codebook in terms of computation
time and lack of training data,
(ii) storing the codebook would require a huge amount of memory, and
(iii) encoding vectors and (iv) pre-computing $f(c_k)$ for all $k$ would be too slow.

\paragraph{Product Quantization (PQ).}
J\'egou et al.\ \cite{Jegou11a} and subsequent related methods \cite{Chen10,Ge13,Norouzi13,Babenko14,Babenko15},
follow the k-means approach but address
its deficiencies by means of a ``product vocabulary'' --
a vector is split into $M$ non-overlapping block where each block is quantized
independently with its own k-means-based quantizer.
The compressed representation of the vector is now the $M$-tuple of IDs
of the closest codewords for each block:
$[a_1(x^{(1)}), a_2(x^{(2)}), \dots, a_M(x^{(M)})]$,
where $x^{(j)}$ be the $j$-th block of vector $x$
and $a_j(x^{(j)})=\argmin_k\sqd{x^{(j)}}{c^{(j)}_k}$.

The size of the quantized representation is $M\log_2K$ and the number of
distinct vectors which can be obtained using this coding scheme is $K^M$,
while the number of codewords which should be learnt is only $M \times K$.
PQ makes it possible to use larger codes than the naive k-means method,
e.g.\ to achieve a 64-bit code one can use $M=8$ and $K=256$, requiring only
$M \times K=2048$ codewords to be learnt from the data, as opposed to k-means
which would require $1.8 \times 10^{19}$.

On the other hand, due to the structure imposed onto the ``product vocabulary''.
it is not possible any more to evaluate arbitrary functions $f(x)$ in
the same manner as before. However, many useful functions are decomposable
over blocks of the input vector and can therefore be evaluated with PQ,
with notable examples being the scalar product and the squared Euclidean distance.
It is possible to estimate the
scalar product between a query vector $q$ and any quantized vector as follows:
\begin{equation}
q^Tx = \sum_{j=1}^M {q^{(j)}}^T x^{(j)}
 \approx \sum_{j=1}^M {q^{(j)}}^T \hat{x}^{(j)}
 = \sum_{j=1}^M {q^{(j)}}^T c^{(j)}_{a_j(x^{(j)})}
\end{equation}
Similarly to the k-means case, for a new query vector $q$, one can pre-compute
the scalar products for each block and each codeword of that block
(done in $O(MK)$) and store them in a lookup table.
Then, it is possible to estimate the scalar product between the query
and all compressed database vectors by simply summing up the scalar product
contributions of each block (done in $O(M)$ per database vector).
A completely analogous strategy can be used to estimate the squared Euclidean
distances as it also decomposes into a sum of per-block squared Euclidean
distances.

The original paper~\cite{Jegou11a} analyzed the effect of product quantization on pairwise distance estimates and found out that PQ has a shrinking effect thus giving biased estimates. They have also suggested an additive factor that leads to an unbiased estimate; we explore this topic in more detail in \sect{bias}.

\paragraph{Optimized Product Quantization (OPQ).}
Due to the splitting of input vectors into blocks, PQ is not invariant to
rotations of the vector space, i.e.\ different rotations of the space
can yield different compression quality. To address this issue,
Ge et al.\ \cite{Ge13} and Norouzi and Fleet \cite{Norouzi13}
propose two very similar methods, which hereafter we jointly refer to as
OPQ. Namely, OPQ aims at finding the best rotation $R$
such that compressing $Rx$ with PQ yields the smallest distortion;
the quantization and rotation are learnt from training vectors by
alternating between learning PQ in a fixed rotated space
with learning the best rotation under the fixed PQ.
The computation of the scalar product or the squared Euclidean distance between a query vector and the compressed database
is the same as above for PQ, apart
from a simple preprocessing step where each vector is rotated by $R$.

\section{The approach}
\label{sect:pairq}

In general, we consider squared distances and scalar products between uncompressed query vectors $q \in R^n$ and compressed dataset vectors $x \in R^n$. We assume that these two vectors come from the two distributions $\Q$ and $\X$, and we assume that at training time we are given datasets of vectors $\{q_i\} \sim \Q$ and $\{x_i\} \sim \X$ coming from these distributions. Note, that we do not assume any similarity between $\Q$ and $\X$.

We now derive the transformations that allows to apply standard quantization methods in a way that the expected distortions of pairwise relations between $q$ and $x$ are minimized. We start with the scalar products and then extend the construction to squared Euclidean distances.

\subsection{Scalar products}
\label{sect:method:dot}

We are interested in efficiently computing the scalar product $q^T x_i$
for all $i$, and to achieve this,
we seek to obtain the approximate representations $\{\hat{x}_i\}$
for the dataset vectors, which will facilitate the efficient approximate
scalar product estimation. Given a training set of query vectors $\{q_i\}$,
the task is to minimize the square loss between the real scalar products
and the scalar product estimates:
\begin{equation}
L_s = \sum_i \sum_j (q_i^T x_j - q_i^T \hat{x}_j)^2
  = \sum_i \sum_j (q_i^T (x_j - \hat{x}_j))^2
\label{eq:scalLoss}
\end{equation}
Further manipulations reveal:
\begin{equation}
L_s = \sum_i \sum_j (x_j - \hat{x}_j)^T q_i q_i^T (x_j - \hat{x}_j)
  = \sum_j (x_j - \hat{x}_j)^T (\sum_i q_i q_i^T) (x_j - \hat{x}_j)
\end{equation}
Defining $G=\sum_i q_i q_i^T$ gives:
\begin{equation}
L_s = \sum_j (x_j - \hat{x}_j)^T G (x_j - \hat{x}_j)
\end{equation}
Note that $G$ is a positive semi-definite matrix as it is a sum of positive
semi-definite matrices $q_i q_i^T$, so it is possible to use SVD to
decompose it into $C$ such that $G=C^T C$.
One can now define a mapping of vectors $x$ into $z$ by $z=Cx$,
and $\hat{z}=C\hat{x}$, yielding the loss:
\begin{equation}
L_s = \sum_j (x_j - \hat{x}_j)^T C^T C (x_j - \hat{x}_j)
  = \sum_j (z_j - \hat{z}_j)^T (z_j - \hat{z}_j)
  = \sum_j \sqd{z_j}{\hat{z}_j}
\label{eq:compressionLoss}
\end{equation}
By means of the aforementioned mapping, we have reduced the original problem
to the standard problem of vector quantization (c.f.\ \eq{quantLoss})
in the transformed $z$-space,
where the goal is to minimize mean squared error between the original
vectors and the vector reconstructions.
Any vector quantization method can be used here, and we employ
Optimized Product Quantization (c.f.\ \sect{opq} for a review).

At query time, to estimate the scalar product between a query $q$
and a database vector $x$, we simply reverse the above order of operations,
i.e.\ the reconstructed $\hat{x}$ are obtained from $\hat{z}$
by inverting the transformation $C$:
\begin{equation}
q^T x \approx q^T \hat{x} = q^T C^{-1} \hat{z},
\label{eq:method:dot}
\end{equation}
where $C^{-1}$ is the (pseudo)inverse of $C$.

Note that at query time it is not actually necessary to explicitly
reconstruct $\hat{z}$ as OPQ can efficiently compute a scalar product
between an uncompressed vector (here $\left(C^{-1}\right)^{T}q$) and a quantized vector
(compressed representation of $\hat{z}$) using additions of values from
a lookup table; for more details on OPQ, recall \sect{opq}.

\subsection{Squared Euclidean distances}
\label{sect:method:sqd}

Analogously to the scalar product preservation task from the previous section,
here we consider the task of efficiently estimating the squared Euclidean distances
between the query and the database vectors $\sqd{q}{x_i}$.
We seek to obtain the approximate database vector representations $\{\hat{x}_i\}$
to minimize the square loss between real and estimated squared distances:
\begin{equation}
L_d = \sum_i \sum_j \left( \sqd{q_i}{x_j} - \sqd{q_i}{\hat{x}_j} \right)^2
\label{eq:euclDistLoss}
\end{equation}
Expanding the inner squares gives:
\begin{equation}
L_d = \sum_i \sum_j \left( -2q_i^Tx_j + \sqn{x_j} +2q_i^T\hat{x}_j - \sqn{\hat{x}_j} \right)^2
\end{equation}
Consider the following definition of vectors $g$ and $y$:
$g=[-2q^T,1]^T$ and $y=[x^T, \sqn{x}]^T$,
simplifying the expression for the loss to:
\begin{equation}
L_d = \sum_i \sum_j \left( g_i^Ty_j - g_i^T\hat{y}_j \right)^2
\label{eq:distToScal}
\end{equation}
Comparing \eq{distToScal} to \eq{scalLoss} it is clear that the Euclidean
distance preservation problem has been reduced to the scalar product
preservation problem. Namely, the $G$-matrix is formed as
$G=\sum_i g_i g_i^T$ and SVD is applied to obtain the
transformation matrix $C$ (a matrix square root of $G$).
The $z$-vectors are then formed as $z=Cy=C [x^T, \sqn{x}]^T$ and the task
is again reduced to minimizing $L_d= \sum_j \sqd{z_j}{\hat{z}_j}$,
for which we again employ OPQ.

At query time, to estimate the squared distance between a query $q$
and a database vector $x$, we follow the procedure analogous
to the scalar product one:
\begin{equation}
\sqd{q}{x}=\sqn{q}+g^Ty
 \approx \sqn{q}+g^T\hat{y}
 = \sqn{q}+g^T C^{-1} \hat{z}
\end{equation}
\subsection{Estimation bias}
\label{sect:bias}

In this section we show that our methods provide unbiased estimates for the
scalar product and the squared Euclidean distance, while OPQ and other vector quantization methods are
only unbiased when used for the scalar product approximation.

For a given query vector $q$, we are interested in the difference in
the expected values of the scalar function $f(q,x)$ and its estimator
$f(q,\hat{x})$ which uses the approximate version $\hat{x}$ of $x$,
where $x \sim \X$, i.e.:
\begin{equation}
\mathrm{\mathbb E_{x \sim \X}}\left(f(q,x)-f(q,\hat{x})\right)
\equiv \E{f(q,x)-f(q,\hat{x})}
\end{equation}
We start by investigating the OPQ estimation bias, and then use the derived results
to analyze our Pairwise Quantization.

\subsubsection{Optimized Product Quantization}

Since the computations of the scalar product and the squared Euclidean distance
decompose across quantization blocks for (Optimized) Product Quantization
(i.e.\ the scalar product estimate is the sum over scalar product estimates
over all quantization blocks), it is sufficient to consider the single
quantization block case ($M=1$) without loss of generality.
For this setting, learning the OPQ codebook $\{c_i\}$ corresponds to performing k-means
on all training vectors, while the quantization is performed by simply assigning
the vector to the nearest cluster centre: $a(x)=\argmin_k\sqd{x}{c_k}$.
The approximation $\hat{x}$ of $x$ is then $\hat{x}=c_{a(x)}$.
To evaluate the estimation bias, if will be beneficial to do so on a per-cluster basis,
i.e.\ the bias will be computed for all $x$ assigned to a particular cluster
centre $k$:
\begin{equation}
\E{ f(q,x) - f(q,\hat{x})~ | ~ a(x)=k } \equiv \Ek{ f(q,x) - f(q,\hat{x}) }
 = \Ek{ f(q,x) } - f(q,\hat{x})
\end{equation}
Then the overall bias can be computed as:
\begin{equation}
\E{ f(q,x) - f(q,\hat{x}) } = \sum_k P(a(x)=k) \left( \Ek{ f(q,x) } - f(q,\hat{x}) \right),
\end{equation}
which is a weighted sum of the per-cluster biases.

\paragraph{Scalar product.}
The function of interest is $f(q,x)=q^Tx$. As $q$ is a constant independent of $x$ it
straightforwardly follows that:
\begin{equation}
\Ek{ f(q,x) } - f(q,\hat{x})
 = \Ek{q^Tx}-q^T\hat{x}
 = q^T\Ek{x}-q^T c_k
 = 0,
\end{equation}
where the last equality comes from the fact that
$\Ek{x}=c_{a(x)}=c_k$ from the definition of the k-means algorithm
(i.e.\ the cluster centre is the mean of the vectors assigned to it).
As all per-cluster biases are equal to zero, and the overall bias is a weighted
sum of per-cluster biases, the overall bias is zero as well, so
OPQ produces an unbiased estimate of the scalar product.

\paragraph{Squared Euclidean distance.}
Investigation of the bias of squared Euclidean distance estimates of (O)PQ
has already been performed in the original PQ paper \cite{Jegou11a},
but we include it here for completeness.

The function of interest is $f(q,x)=\sqd{q}{x}$. Using again the fact
that $\Ek{x}=c_k$, the expected squared distance
for a given cluster $k$ is:
\begin{eqnarray}
\Ek{\sqd{q}{x}} &=& \sqn{q}-2q^T\Ek{x}+\Ek{\sqn{x}} = \sqn{q}-2q^Tc_k+\Ek{\sqn{x}} \\
 &=& \sqd{q}{c_k} - \sqn{c_k} + \Ek{\sqn{x}} \\
 &=& \sqd{q}{c_k} - \sqn{c_k} + \Ek{\sqn{x-c_k+c_k}} \\
 &=& \sqd{q}{c_k} - \sqn{c_k} + \Ek{\sqd{x}{c_k}-2c_k^T(x-c_k)+\sqn{c_k}} \\
 &=& \sqd{q}{c_k} + \Ek{\sqd{x}{c_k}} - 2c_k^T(\Ek{x}-c_k) \\
 &=& \sqd{q}{c_k} + \Ek{\sqd{x}{c_k}}
\label{eq:biasCorrection}
\end{eqnarray}
As $\sqd{q}{c_k}$ is the estimate of the squared distance, the per-cluster bias
is therefore $\Ek{\sqd{x}{c_k}}$. This term corresponds to the
mean squared error (MSE) and is in general larger than zero.
Since the overall bias is a weighted sum of per-cluster biases, where the
weights are all non-negative and in general larger than zero,
the overall bias is larger than zero.
Therefore, OPQ produces a biased estimate for the squared Euclidean distance,
and it on average provides an underestimate.

For the case where there are more than one
subquantizers ($M>1$), since the squared distance estimates are additive, it is easy
to show that the bias correction term is equal to the sum of the individual
MSE's \cite{Jegou11a}.

\paragraph{Generalization to other methods.}
Other popular methods for vector quantization such as
Residual Vector Quantization (RVQ) \cite{Chen10},
Additive Quantization (AQ) \cite{Babenko14},
Tree Quantization (TQ) \cite{Babenko15}, and
Composite Quantization (CQ) \cite{Zhang14}
are all equivalent to OPQ when the number of subquantizers is $M=1$
and therefore in general yield biased squared Euclidean distance estimates.
Regarding the scalar product estimates --
RVQ is unbiased, which can be proven in the same way as for OPQ,
while for the more involved methods AQ, TQ and CQ
it is not clear whether they are biased or not.

\subsubsection{Pairwise Quantization}

\paragraph{Scalar product.}
Recall from \sect{method:dot} and \eq{method:dot} that $z=Cx$
and that therefore the scalar product is $q^Tx=q^TC^{-1}z$.
Defining $r=C^{-T}q$, the estimator bias is:
\begin{equation}
\E{q^Tx-q^T\hat{x}}=\E{ q^TC^{-1}z - q^TC^{-1}\hat{z} } = \E{ r^Tz - r^T\hat{z} }
\end{equation}
As the right hand side corresponds to the bias of the scalar product estimates
in the z-space, the bias of our method depends on the underlying method we
choose to use in the z-space. Since we use OPQ which, as shown earlier,
yields unbiased estimates of the scalar product, our method is also unbiased.

\paragraph{Squared Euclidean distance.}
As shown in \sect{method:sqd}, by introducing new variables
$g=[-2q^T,1]^T$ and $y=[x^T,\sqn{x}]^T$, the squared Euclidean distance
$\sqd{q}{x}$ can be computed as a sum of a constant $\sqn{q}$ and
a scalar product $g^Ty$. Since our method provides unbiased estimates
of the scalar product, it is clear that the squared Euclidean distance
estimates are unbiased as well.
Note that this is different from OPQ which yields biased
estimators.

\section{Applications and experiments}

Now we demonstrate the advantage of Pairwise Quantization (PairQ) on real datasets from three different applications. In all experiments below we use OPQ to perform compression \eq{compressionLoss} and also use OPQ as the main baseline for comparison unless stated otherwise. For both OPQ and PairQ we use codebooks with $K=256$ codewords which is a standard choice in most previous works.

\subsection{Text-to-image retrieval}

The problem of text-to-image retrieval arises in image search engines that rank images based on their relevance to a user text query. The advanced methods for this problem typically use multimodal deep neural networks that learn common representations for both text and images  capturing the semantics of both modalities~\cite{TextToImage15}. Given such representations, the relevance of a particular image to a particular text query is measured as a cosine similarity between them, i.e.\ the scalar product of the L2-normalized features.

The retrieval quality can be improved by using images metadata (e.g. tags, EXIF, etc.). Therefore, modern search engines combine the relevance score provided by the deep representations with features extracted from the metadata via a higher-level ranker that re-orders images based on multiple sources of information. For large image databases, exact evaluation of all cosine distances for a given query is expensive, hence approximate evaluation is used in practice. Clearly, it is important to minimize the corruption of cosine distance values in order to provide good features to the higher-level ranker. We compare OPQ and PairQ for this problem and measure the average squared error of cosine distance approximations for both methods.

We followed the protocol described in \cite{TextToImage15} and trained a multimodal deep network on a set of text-image pairs. The train dataset was collected by ourselves via querying an image search engine with $1300$ popular text queries. For each query we used the top ten images as positives (their relevance to the query equals one) and random images from the web as negatives (their relevance to the query equals zero). After the network was learnt, we computed the representations for $579067$ images and $33235$ text queries. The dimensionality of the representations was set to $100$.
The dataset is available upon request\footnote{%
Please contact Artem Babenko at \texttt{artem.babenko@phystech.edu}}.

In this experiment we used $300000$ image representations for learning OPQ and PairQ codebooks as well as the rotation matrices inside OPQ.  The other $279067$ image representations were used for performance evaluation. To learn the transformation $C$ in PairQ we used $25000$ text representations and the remaining text representations were used only for the performance evaluation of both methods.

\begin{table}
\caption{Mean squared approximation error of cosine distances between image and text representations from the multimodal deep network. For the whole range of compression ratios PairQ provides more accurate compression compared to OPQ. Overall, PairQ allows to reduce approximation error by upto $43\%$.}
\label{tab:retrieval}
\centering
\addtolength{\tabcolsep}{3pt}
\begin{tabular}{|c|c|c|c|c|c|}
\hline
Bytes per vector & 5 & 10 & 20 & 25 & 50\\
\hline
Compression ratio & 80 & 40 & 20 & 16 & 8\\
\hline
OPQ error, $10^{-3}$ & 2.127 & 1.234 & 0.508 & 0.356 & 0.056\\
\hline
PairQ error, $10^{-3}$ & {\bf 1.764} & {\bf 0.961} & {\bf 0.329} & {\bf 0.203} & {\bf 0.045}\\
\hline
Error reduction w.r.t OPQ & 17\% & 22\% & 35\% & 43\% & 20\%\\
\hline
\end{tabular}
\end{table}

The results of comparison are presented in \tab{retrieval}. We used different numbers of codebooks for both OPQ and PairQ thus providing different compression ratios, corresponding to different memory and time budgets. \tab{retrieval} demonstrates that PairQ outperforms OPQ over the whole range of compression ratios, and reduces OPQ error by upto $43\%$.

\subsection{Recommendation systems}

Recommendation systems have become ubiquitous on e-commerce sites where the task is to suggest items to users according to their personal preferences. This problem is typically solved by collaborative filtering (CF) methods that perform a factorization of the user ratings matrix thus producing latent vector representations for each user $u$ and each item $i$. Then the relevance of an item to a specific user can be obtained as the value of the scalar product $u^Ti$. The alternative approach to the personalization problem is to use content-based filtering (CBF) that is based on given descriptions of users and items (e.g.\ tags, clicks, user history, etc.). 

The modern systems typically combine both approaches and use the outputs of CF and CBF as features for a higher-level recommender \cite{GunawardanaM09}. Hence to produce recommendations for a user with a latent representation $u$, the scalar products $\{u^Ti_k\}_{k=1}^{N}$ should be computed, where $\{i_1,\dots,i_N\}$ is the items database. For large $N$, the exact evaluation of scalar products can be computationally expensive and efficient approximate methods should be used. Along with efficiency, approximate methods should provide accurate reconstructions of scalar product values that will be then used by the higher-level recommender. We compare OPQ and PairQ for this problem of approximate scalar product evaluation and use the average squared error of approximation as the performance measure.

We perform the comparison on the well-known MovieLens-20M dataset \cite{MovieLens}. Latent user and item representations of dimensionality $150$ were obtained by the standard PureSVD procedure \cite{CremonesiKT10}. Overall this dataset contains $138493$ user vectors and $26744$ item vectors. We used the first $100000$ user vectors to learn the transformation $C$ and the other user vectors to evaluate the approximation quality. 

\begin{table}
\caption{Average squared approximation error of scalar products between user vectors and item vectors from MovieLens dataset. For the whole range of compression ratios PairQ provides more accurate compression comparing to OPQ. Overall, PairQ allows to reduce approximation error by upto $42\%$.}
\label{tab:movielens}
\centering
\addtolength{\tabcolsep}{3pt}
\begin{tabular}{|c|c|c|c|c|c|}
\hline
Bytes per vector & 5 & 10 & 15 & 25 & 50\\
\hline
Compression ratio & 120 & 60 & 40 & 24 & 12\\
\hline
OPQ error, $10^{-3}$ & 7.570 & 6.952 & 5.828 & 4.432 & 1.793\\
\hline
PairQ error, $10^{-3}$ & {\bf 6.464} & {\bf 4.601} & {\bf 3.719} & {\bf 2.529} & {\bf 1.297}\\
\hline
Error reduction w.r.t OPQ & 15\% & 34\% & 36\% & 42\% & 28\%\\
\hline
\end{tabular}
\vspace{-4mm}
\end{table}

We provide the results of comparison for different compression ratios in \tab{movielens}. It shows that PairQ provides significantly more accurate scalar product approximations than OPQ. For instance, the usage of PairQ instead of OPQ allows to reduce the squared approximation error by $42\%$.

\subsection{Preserving Euclidean distances}

Many machine learning algorithms require efficient evaluation of Euclidean distances between a given query and a large number of database vectors. Several examples are large-scale SVMs with RBF kernel, kernelized regression and non-parametric density estimation. We show that Pairwise Quantization provides more accurate reconstructions of Euclidean distances comparing to OPQ for the same runtime and memory costs.

The experiments were performed on three datasets. The \textbf{SIFT1M} dataset~\cite{Jegou11a} contains one million of $128$-dimensional SIFT descriptors \cite{Lowe04} along with $10,000$ predefined queries. The \textbf{GIST1M} dataset~\cite{Jegou11a} contains one million of $960$-dimensional GIST descriptors \cite{Oliva01} and $1,000$ queries. The \textbf{DEEP1M} dataset~\cite{Babenko15} contains one million of $256$-dimensional deep descriptors and $1,000$ queries. Deep descriptors are formed as PCA-compressed outputs from a fully-connected layer of a CNN pretrained on the Imagenet dataset \cite{Deng09}.

\begin{figure}
\noindent
\centering
\begin{tabular}{ccc}
\includegraphics[width=4.3cm]{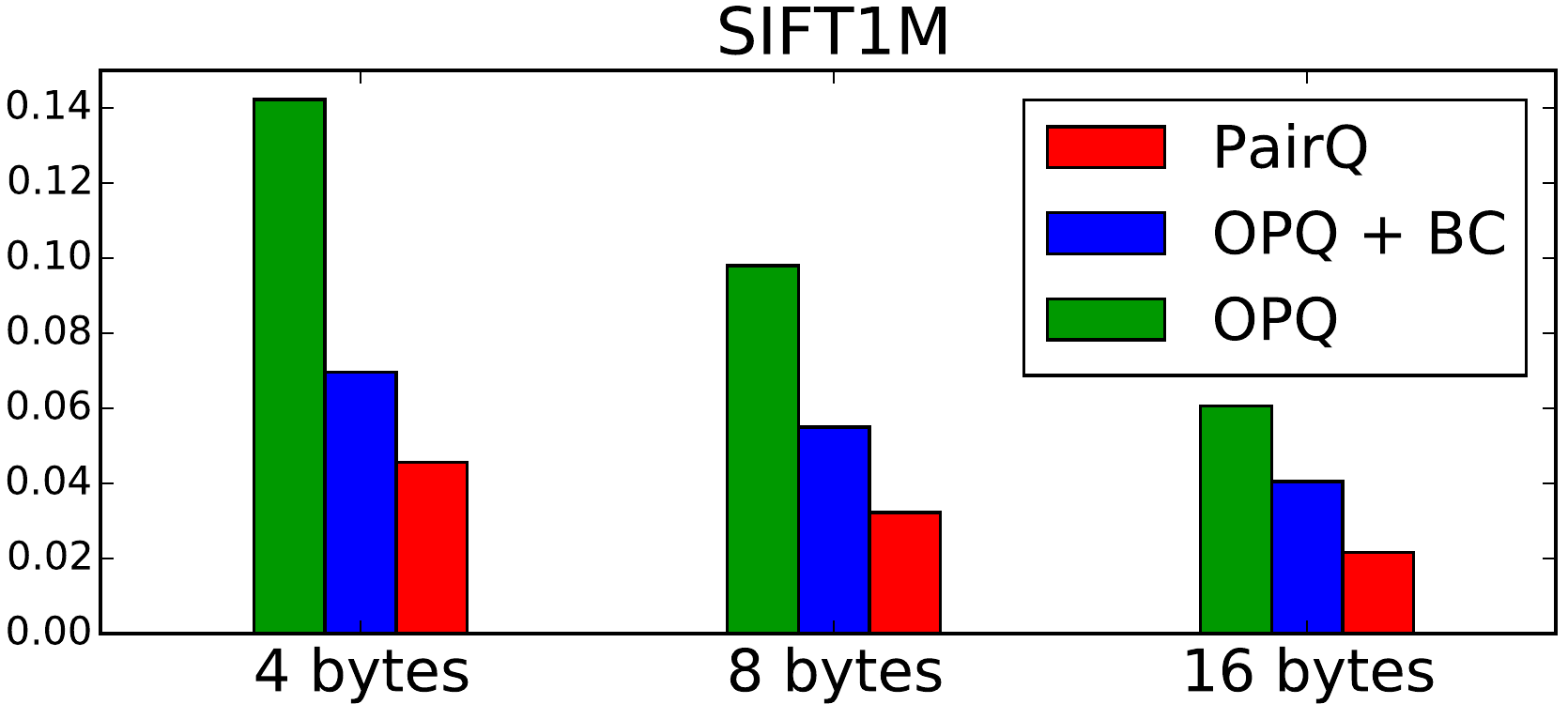}&
\includegraphics[width=4.3cm]{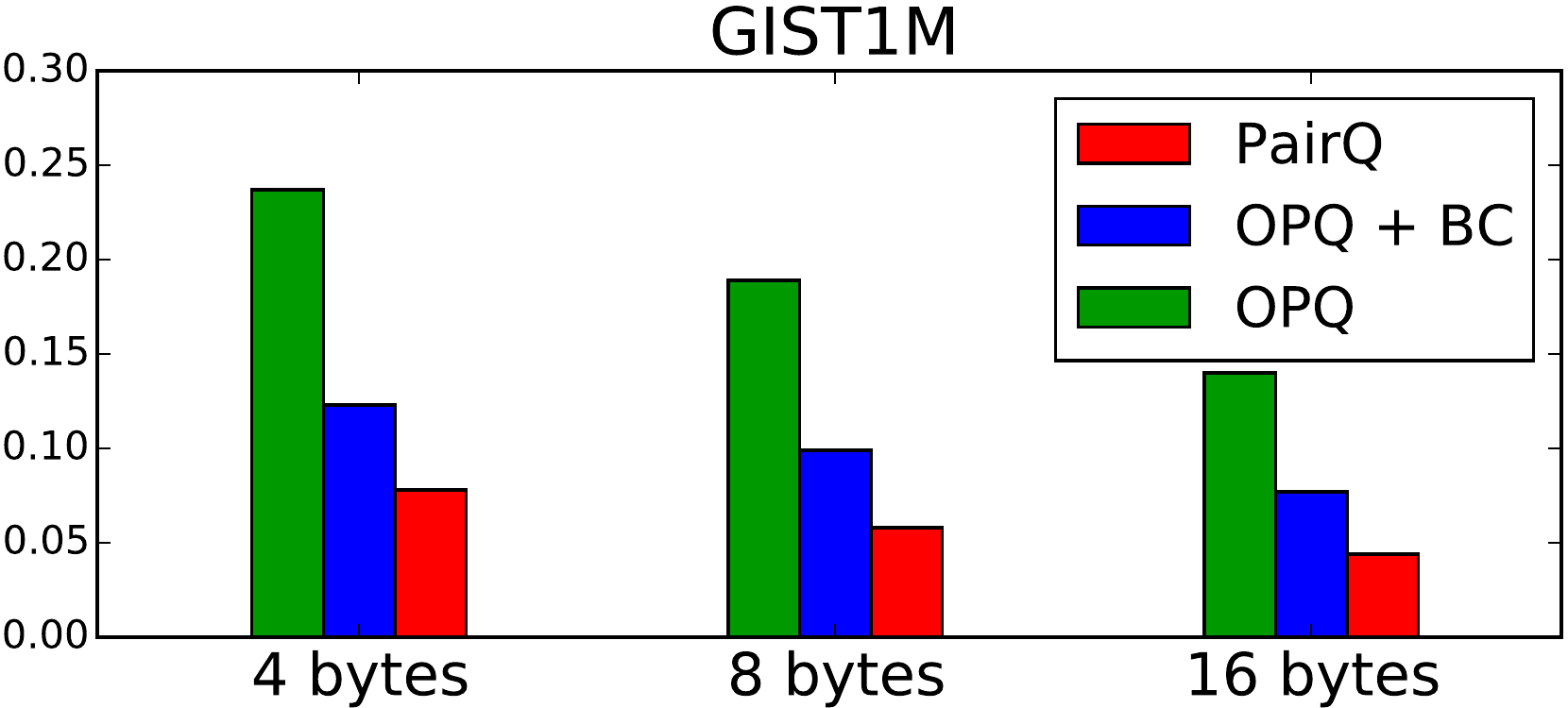}&
\includegraphics[width=4.3cm]{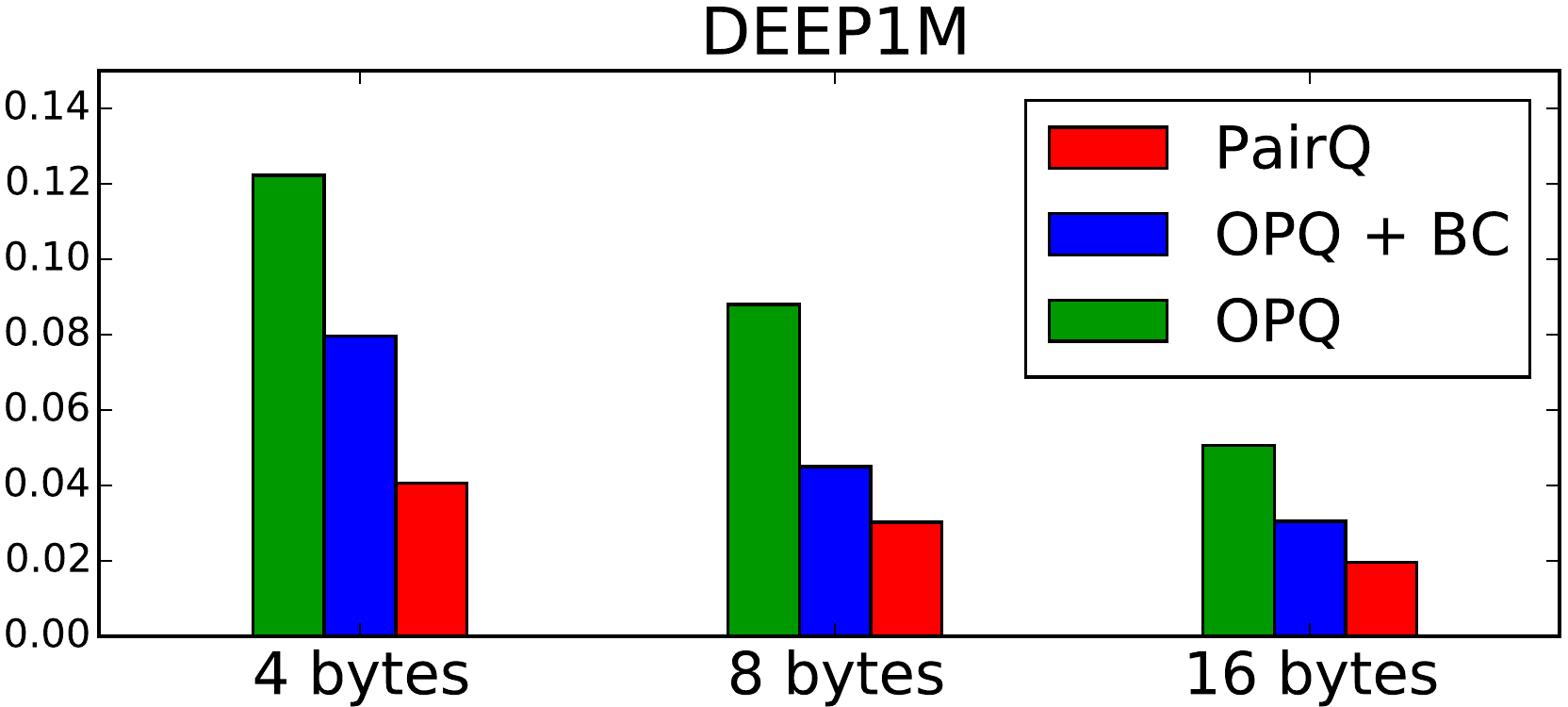}\\
\includegraphics[width=4.3cm]{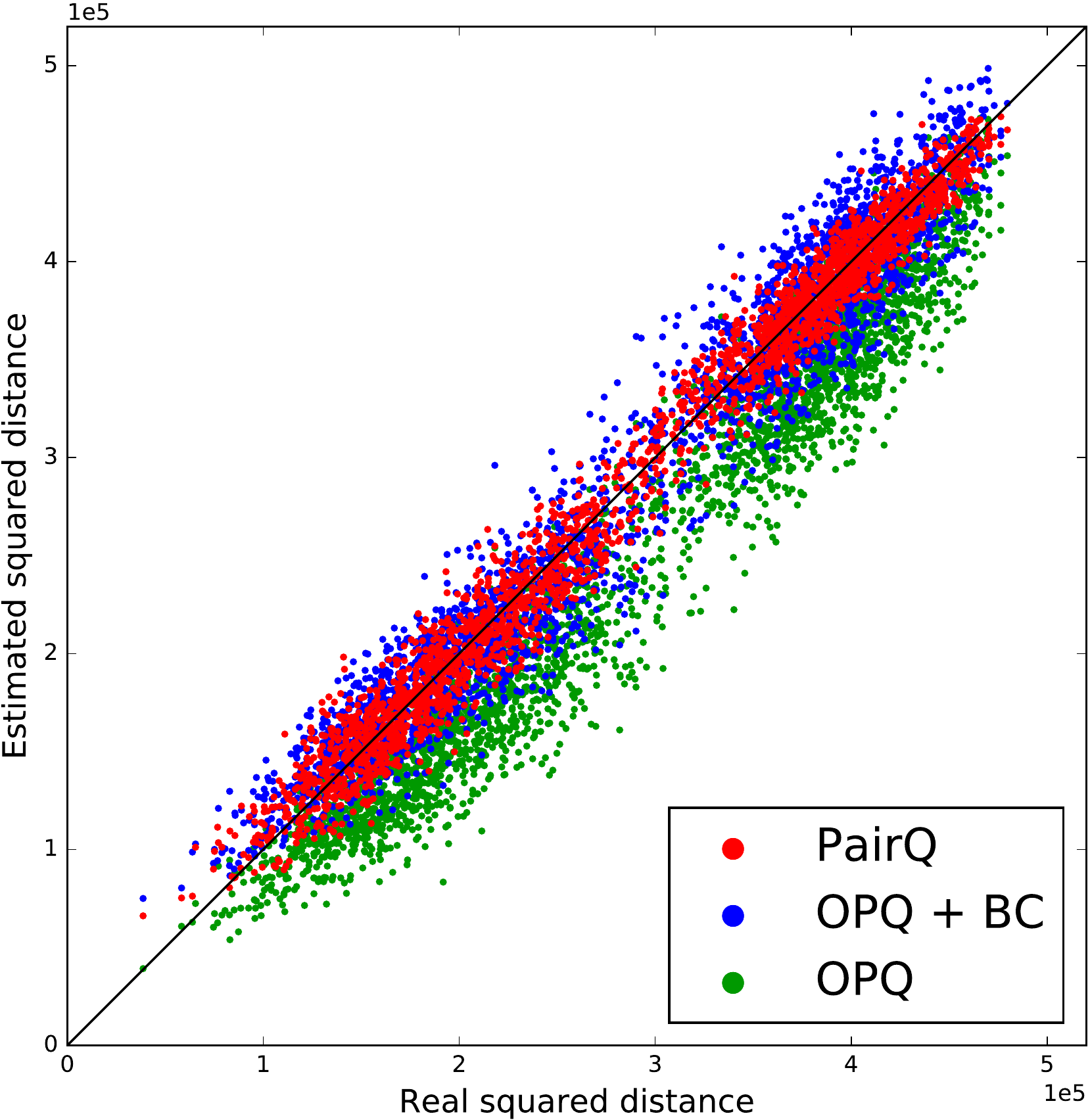}&
\includegraphics[width=4.4cm]{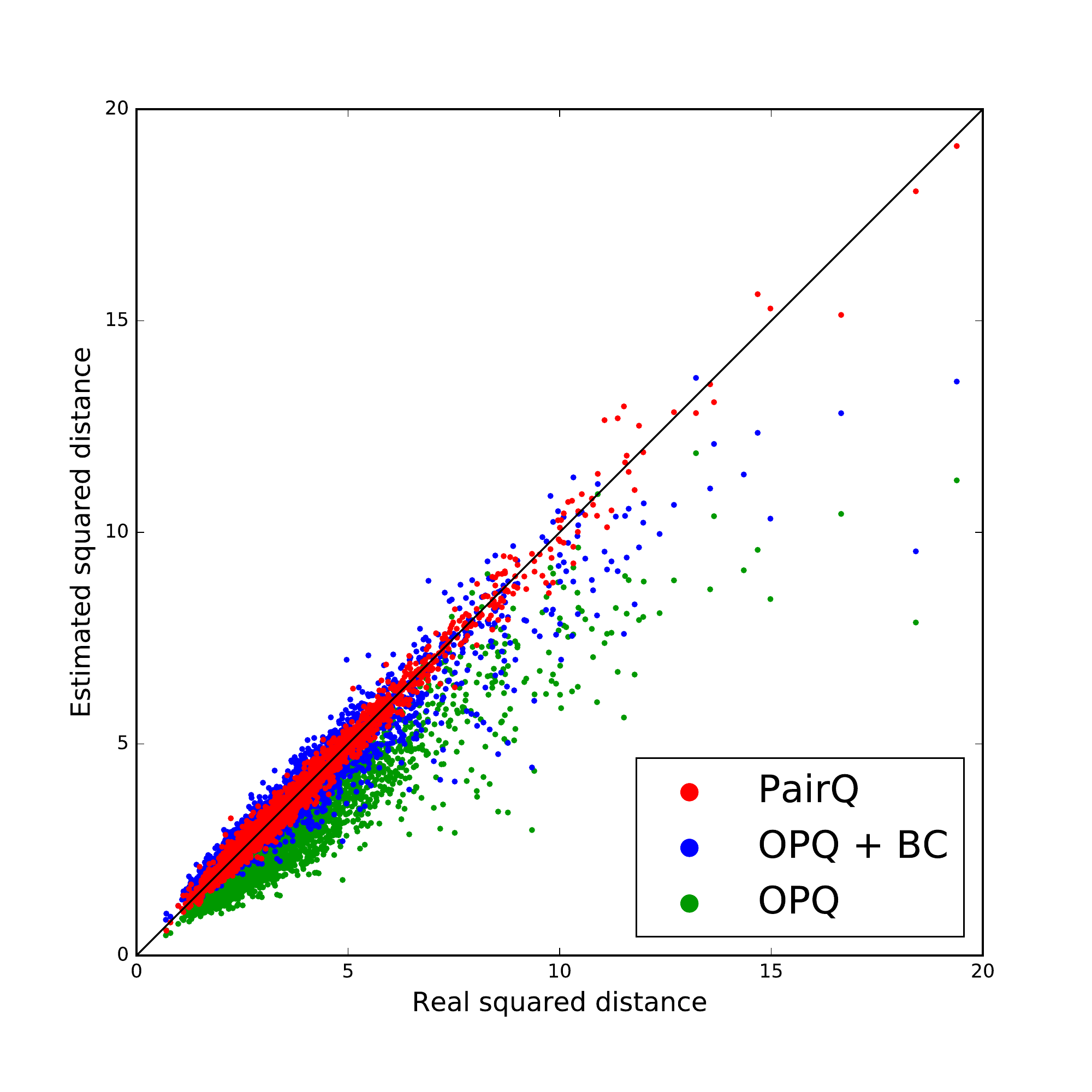}&
\includegraphics[width=4.35cm]{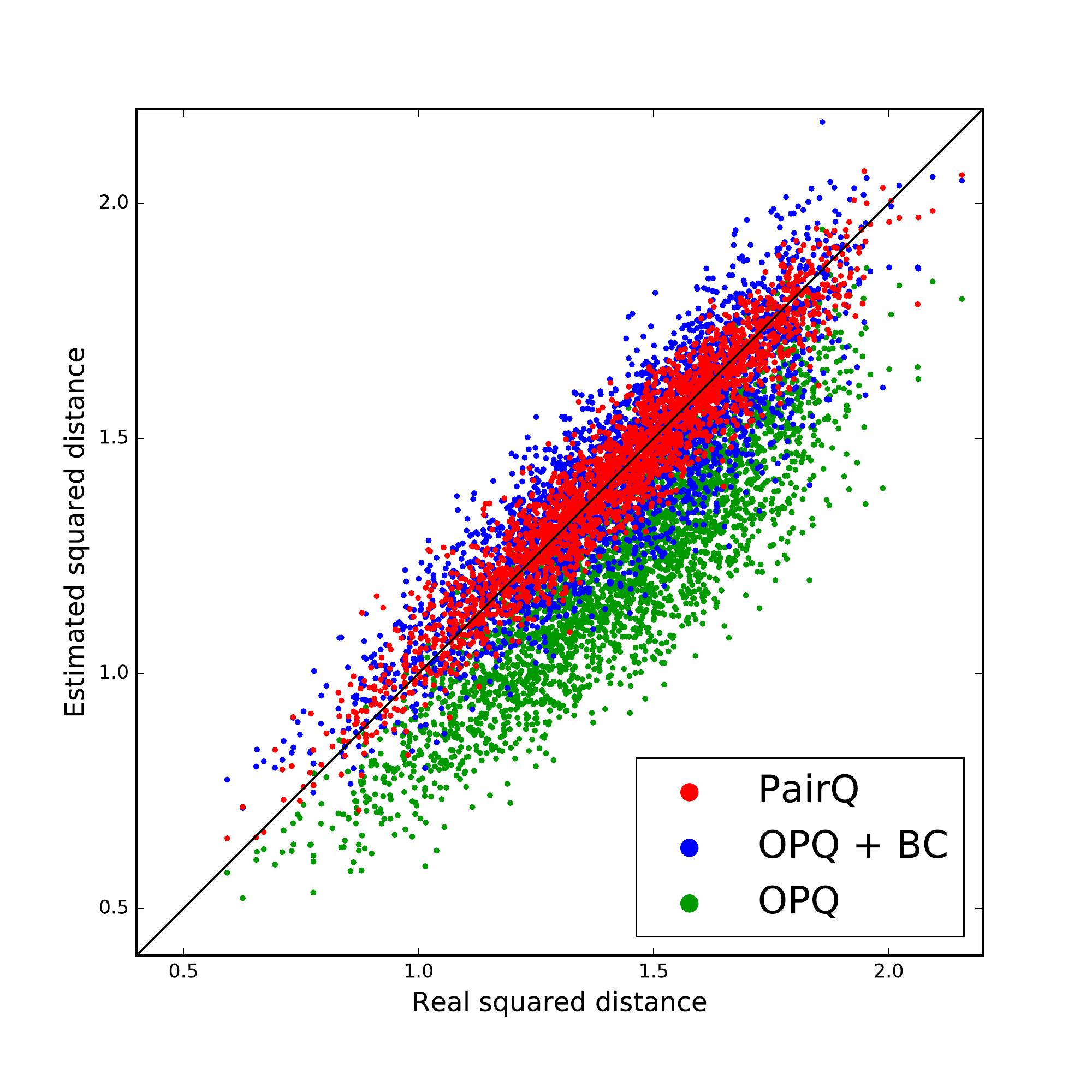}
\end{tabular}
\vspace{-2mm}
\caption{Top row shows mean relative errors of Euclidean distances approximation for different methods and memory budgets. Bottom row contain scatterplots for the case of $M=4$ with scatterpoints corresponding to various (query, dataset point) pairs, and scatterpoint coordinates set to be (real distance, estimated distance) with the diagonal line containing perfect estimates. OPQ, OPQ with bias correction (OPQ+BC) and PairQ are compared. On all datasets PairQ outperforms OPQ and OPQ+BC by a large margin.} 
\label{fig:distances}
\vspace{-4mm}
\end{figure}

Along with OPQ as the first baseline, we also include OPQ with bias correction (OPQ+BC) as the second baseline. The bias correction term is computed as described in \eq{biasCorrection} and provides more accurate distance estimation as we show below. For all datasets, the codebooks for OPQ and PairQ and the transformation $C$ in PairQ are learnt on the provided hold-out sets. 

For each query $q$ and each database vector $x$ we evaluate a relative approximation error of the Euclidean distance produced by OPQ, OPQ+BC and PairQ. \fig{distances} (top row) shows relative errors averaged over all possible $(q,x)$ pairs and for different memory budgets per vector. For all datasets PairQ significantly outperforms OPQ and OPQ with bias correction. Note, PairQ requires several times less memory per vector to provide the same accuracy, e.g.\ the performance of PairQ with four bytes per vector is on par with the performance of OPQ with 16 bytes per vector.

We also experimented with the usage of PairQ for the approximate nearest neighbor search problem. For this application PairQ provided inferior results on all datasets comparing to OPQ. The reason for the low performance of PairQ here is that the functional \eq{euclDistLoss} puts too much emphasis on the distant point pairs while only the close pairs are important for ANN search. 
A similar effect (better distance estimates but worse ANN performance) was noticed in \cite{Jegou11a} for their bias correction procedure.

\section{Summary}

We have proposed a new approach for quantization methods that proceeds by minimizing distortions of pairwise relations on the training set. This is in contrast to previous works that optimize the reconstruction error of individual points. We develop a simple technique based on linear transformation that allows to reduce the task of minimizing pairwise distortions to the task of minimizing the reconstruction error in the transformed space. This allows us to adapt previously proposed quantization methods to minimize pairwise distortions directly. 

The experiments confirm that our approach achieves significant reduction in pairwise distortions, when squared distances and scalar product between compressed and uncompressed vectors are considered. We note that beyond retrieval and recommendation systems, the improvement in scalar product estimation accuracy can be useful for learning classifiers and detectors in a large-scale setting \cite{Sanchez11,Vedaldi12}.
Finally, we note that our method is almost straightforwardly applicable beyond quantization, and can turn any learning-based lossy compression method that optimizes the reconstruction error into a method that minimizes expected pairwise distortion. We plan to investigate this ability further in future work.

\paragraph{Acknowledgements.}
Relja Arandjelovi\'c is supported by the ERC grant LEAP (no.\ 336845).
Victor Lempitsky is supported by the Russian Ministry of Science and Education grant RFMEFI61516X0003.

\bibliographystyle{ieee}
\bibliography{refs}

\begin{thebibliography}{10}\itemsep=-1pt

\bibitem{Babenko14}
A.~Babenko and V.~Lempitsky.
\newblock Additive quantization for extreme vector compression.
\newblock {\em CVPR}, 2014.

\bibitem{Babenko15}
A.~Babenko and V.~Lempitsky.
\newblock Tree quantization for large-scale similarity search and
  classification.
\newblock {\em CVPR}, 2015.

\bibitem{Chen10}
Y.~Chen, T.~Guan, and C.~Wang.
\newblock Approximate nearest neighbor search by residual vector quantization.
\newblock {\em Sensors}, 10(12):11259--11273, 2010.

\bibitem{CremonesiKT10}
P.~Cremonesi, Y.~Koren, and R.~Turrin.
\newblock Performance of recommender algorithms on top-n recommendation tasks.
\newblock {\em Proceedings of the 2010 {ACM} Conference on Recommender Systems,
  RecSys 2010, Barcelona, Spain, September 26-30, 2010}, pp. 39--46, 2010.

\bibitem{Deng09}
J.~Deng, W.~Dong, R.~Socher, L.-J. Li, K.~Li, and L.~Fei-Fei.
\newblock Imagenet: A large-scale hierarchical image database.
\newblock {\em CVPR}, 2009.

\bibitem{TextToImage15}
H.~Fang, S.~Gupta, F.~N. Iandola, R.~K. Srivastava, L.~Deng, P.~Doll{\'{a}}r,
  J.~Gao, X.~He, M.~Mitchell, J.~C. Platt, C.~L. Zitnick, and G.~Zweig.
\newblock From captions to visual concepts and back.
\newblock {\em {IEEE} Conference on Computer Vision and Pattern Recognition,
  {CVPR} 2015, Boston, MA, USA, June 7-12, 2015}, pp. 1473--1482, 2015.

\bibitem{Ge13}
T.~Ge, K.~He, Q.~Ke, and J.~Sun.
\newblock Optimized product quantization for approximate nearest neighbor
  search.
\newblock {\em CVPR}, 2013.

\bibitem{GunawardanaM09}
A.~Gunawardana and C.~Meek.
\newblock A unified approach to building hybrid recommender systems.
\newblock {\em Proceedings of the 2009 {ACM} Conference on Recommender Systems,
  RecSys 2009, New York, NY, USA, October 23-25, 2009}, pp. 117--124, 2009.

\bibitem{MovieLens}
F.~M. Harper and J.~A. Konstan.
\newblock The movielens datasets: History and context.
\newblock {\em TiiS}, 5(4):19, 2016.

\bibitem{Jegou11a}
H.~J{\'e}gou, M.~Douze, and C.~Schmid.
\newblock Product quantization for nearest neighbor search.
\newblock {\em TPAMI}, 33(1), 2011.

\bibitem{Lowe04}
D.~G. Lowe.
\newblock Distinctive image features from scale-invariant keypoints.
\newblock {\em IJCV}, 60(2), 2004.

\bibitem{Norouzi13}
M.~Norouzi and D.~J. Fleet.
\newblock Cartesian k-means.
\newblock {\em CVPR}, 2013.

\bibitem{Oliva01}
A.~Oliva and A.~Torralba.
\newblock Modeling the shape of the scene: A holistic representation of the
  spatial envelope.
\newblock {\em IJCV}, 42(3), 2001.

\bibitem{Sanchez11}
J.~S{\'a}nchez and F.~Perronnin.
\newblock High-dimensional signature compression for large-scale image
  classification.
\newblock {\em CVPR}, pp. 1665--1672, 2011.

\bibitem{Vedaldi12}
A.~Vedaldi and A.~Zisserman.
\newblock Sparse kernel approximations for efficient classification and
  detection.
\newblock {\em CVPR}, pp. 2320--2327, 2012.

\bibitem{Zhang14}
T.~Zhang, C.~Du, and J.~Wang.
\newblock Composite quantization for approximate nearest neighbor search.
\newblock {\em ICML}, 2014.

\end{thebibliography}

\end{document}